# Hybrid Systems Knowledge Representation Using Modelling Environment System Techniques Artificial Intelligence

**Kamran Latif**
**University of Lahore**
*Email: kamranlatif@gmail.com*

*Abstract*
*Knowledge-based or Artificial Intelligence techniques are used increasingly as alternatives to more classical techniques to model ENVIRONMENTAL SYSTEMS. Use of Artificial Intelligence (AI) in environmental modelling has increased with recognition of its potential. In this paper we examine the DIFFERENT TECHNIQUES of Artificial intelligence with profound examples of human perception, learning and reasoning to solve complex problems. However with the increase of complexity better methods are required. Keeping in view of the above some researchers introduced the idea of hybrid mechanism in which two or more methods can be combined which seems to be a positive effort for creating a more complex; advanced and intelligent system which has the capability to in- cooperate human decisions thus driving the landscape changes.*

## Introduction (Expert system)

ES are computer programs which are derived from a branch of CSR called AI; Artificial intelligence ultimate target is to achieve the utmost intelligence by creating computer programs that demonstrate intelligent activities. ES been made in the area of problem solving, concepts and methods for building programs are to trace the reason about problems rather than calculate a solution. The ES term is often used for such programs whose knowledge based contains the knowledge of human expert or gathered by some books or other relevant material which can also be found online if the knowledge engineer or programmer has ensured that the coded program or computer to self gather the information (this information is to be stored in the computer memory in form of symbols patterns) on particular topic or topics.

### 1. Artificial Intelligence Techniques

Knowledge representation techniques are discussed based on which knowledge about different machines and intelligence can be represented accordingly. Knowledge-based or Artificial Intelligence techniques are used increasingly as alternatives to more classical techniques to model environmental systems.

### 2. HYBRID LEARNING SYSTEM

Hybrid system learning methods use theoretical knowledge of a domain and a set of classified examples to develop a method for precisely classifying examples not seen during training. The challenge of hybrid learning system is to use the information provided by one source of information to offset informant missing from the other source. By doing this a hybrid learning system should learn more effectively than systems that use only one information source. The same is further exemplifying in Fig 1.1 Knowledge Based ANN hybrid learning system.

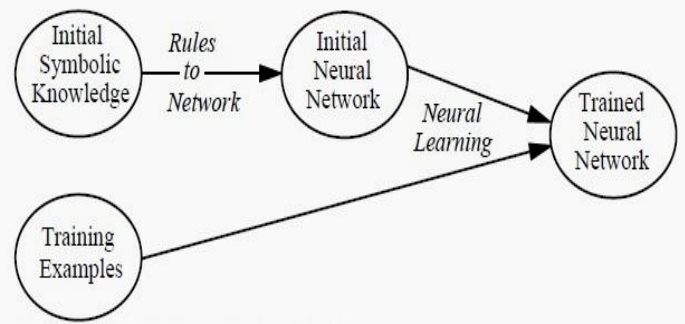

Figure 1.1 Hybird learning System (ANN)

*Further suppose you are trying to teach someone who has never seen a class of objects to recognize members of that class.*

*One approach is to define the category for your student. That is, state a "domain theory"1 that describes how to recognize critical facets of class members and how those interact. Using this domain theory, your student could distinguish between members and nonmembers of the class.*

*A different approach to teaching someone to recognize a class of objects is to show the person lots of examples. As each example is shown, you would tell your student only whether the example is, or is not, a member of the class. After seeing sufficient Examples, your student could classify new examples by comparison to those already seen.*

## 2.1 Hand-Built Classifiers

Hand-built classifiers are non-learning systems (except they are later altered by hand). They simply do what they are told; they do not learn at the knowledge level. Despite their apparent simplicity, such systems simulate many problems for those that build them. Typical hand build classifiers assume that there domain theory is complete; however for most real world tasks totality and accuracy are extremely difficult if not impossible to attain. To make a domain theory as complete and correct as possible, it may be necessary to write thousands of interacting, possibly recursive rules. Use of such rule sets may be intolerably slow. Further such theories can also be difficult to modify

## 2.2 Empirical Learning

Empirical research is a way of gaining knowledge by means of direct and indirect observations or experience; further the evidence can be recorded and analyzed in both manners quantitatively or qualitatively. The empirical learning system requires little theoretical knowledge about the problem domain; instead they require a large library of examples. **According to AD. DE GROOT's Empirical leaning has five Cycle or stages**.

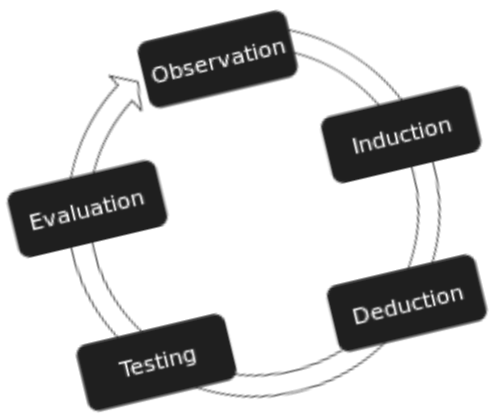

**Figure 1.2 Empirical Cycles**

1. **Observation**: The collecting and organization of empirical facts; forming hypothesis.
2. **Induction**: Formulating hypothesis.
3. **Deduction**: Deducting consequences of hypothesis as testable predictions.
4. **Testing**: Testing the hypothesis with new empirical material.
5. **Evaluation**: Evaluating the outcome of testing

## 3. Artificial Neural Networks ANN

Artificial neural networks (ANNs) show the picture of the human brain as it comes to processes information. Artificial neural networks (ANNs), which form the basis of KBANN, are a particular method for empirical learning. ANNs have proven to be equal, or superior, to other empirical learning systems over a wide range of domains, when evaluated in terms of their generalization ability. ANNs are useful in solving data-intensive problems where the algorithm or rules to solve the problem are unknown or difficult to express. ANNs process information in parallel and are robust to data errors; Further ANNs can be applied to seven categories of problems pattern classification, clustering, function approximation, prediction, optimization, retrieval by content and process control e.t.c

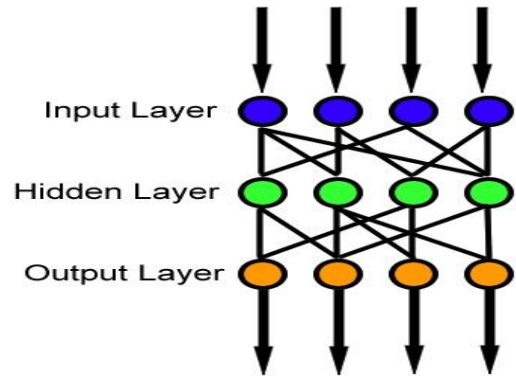

**Figure 1.3 Artificial Neural Networks (Feed Forward)**

## 4. CASE BASED REASONING

Case-based reasoning (CBR) solves a problem by recalling similar past problems. Numerous past cases are needed to adapt their solutions or methods to the new problem CBR recognizes that problems are easier to solve by repeated attempts, accruing learning. It involves four steps as also shown in Figure 1.4

1. **Retrieve** the most relevant past cases from the database;
2. **Use the retrieved case** to produce a solution of the new problem;
3. **Revise the proposed solution** by simulation or test execution; and
4. **Retain the solution** for future use after successful adaptation.

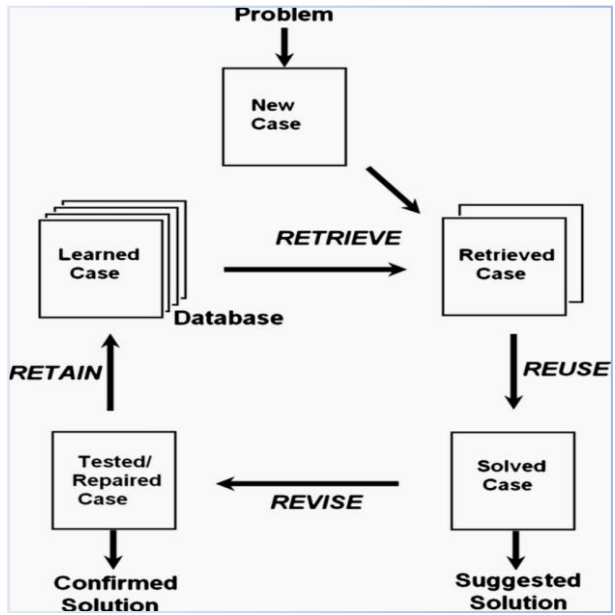

**Figure 1.4 CBR Four step process**

CBR can be employed for diagnosis, prediction, control and planning, Environmental applications include decision and support, managing water treatment plants, monitoring Air quality and weather predictions e.t.c

## 5. RULE BASED SYSTEM

Rule-based systems (RBS) solve problems by rules derived from expert knowledge. The rules have condition and action parts, if and then and are fed to an inference engine, which has a working memory of information about the Problem, pattern matcher and a rule applier. The pattern matcher refers to the working memory to decide which rules are relevant, then the rule applier chooses what rule to apply. New information created by the action (then-) part of the rule applied is added to the working memory and the match-select-act cycle between working memory and knowledge base repeated until no more relevant rules are found. *RBS are easy to understand, implement and maintain, as knowledge is prescribed in a uniform way, as conditional rules. However, the solutions are generated from established rules and RBS involve no learning. They cannot automatically add or modify rules. So a rule-based system can only be implemented if comprehensive knowledge is available.*

## 6. Fuzzy System

Fuzzy systems (FS) use fuzzy sets to deal with imprecise and incomplete data. In conventional set theory an object is a member of a set or not, but fuzzy set membership takes any value between 0 and 1. Thus fuzzy models can describe vague statements as in natural language. Human reasoning handles unclear or imprecise information. The ability of fuzzy systems to handle such information is one of its main strengths over other AI techniques, although they are mostly easier to understand and apply. *Fuzzy systems have no learning capability or memory. To overcome such limitations, fuzzy modelling is often combined with other techniques to form hybrid systems e.g. with neural networks to form Neuro-fuzzy systems.*

Fuzzy systems handle incomplete or imprecise data in applications including function approximation classification/ Clustering, control and prediction e.t.c

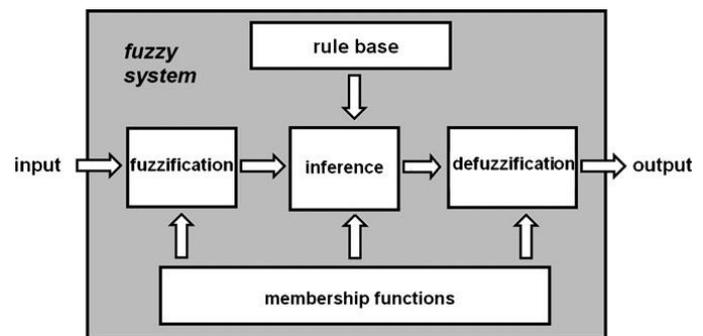

**Figure 1.4 Fuzzy Model Main Components**

## 7. Conclusion

There is a significant gap between the knowledge intensive learning by being asked and told approach of hand build classifier and the VK free approach of empirical learning.

This gap has been tried to be filled up by introducing the hybrid learning methods which use both hand constructed rules and classified examples during learning. Further the suitability of AI techniques for environmental modelling is also case-specific. Complex and poorly understood processes may favour an approach CBR, RBS or ANN's.

CBR requires similar past cases; whereas ANN learns from training cases and captured relationship among data and for problems with well understood process RBS can be applied **Finally the hybrid systems have proven effective on several real world problems**.